\documentclass[runningheads]{llncs}
\usepackage[T1]{fontenc}
\usepackage{graphicx}
\usepackage[utf8]{inputenc} 
\usepackage{amsmath,cite,url,tikz}

\usepackage{graphicx}
\usepackage{color}
\usepackage{booktabs}
\usepackage{array}
\usepackage[separate-uncertainty]{siunitx}
\usepackage[inline]{enumitem}

\usepackage{subcaption}

\usetikzlibrary{fit}
\usetikzlibrary{positioning}

\usepackage{pifont}
\newcommand{\cmark}{\ding{51}}%
\newcommand{\xmark}{\ding{55}}%

\DeclareSIUnit\thousand{k}
\DeclareSIUnit\million{M}

\setlist[description]{font=\normalfont\itshape}

\title{WikiMuTe: A web-sourced dataset of semantic descriptions for music audio}
\author{Benno Weck\inst{1,2} \and
Holger Kirchhoff\inst{1} \and
Peter Grosche\inst{1} \and
Xavier Serra\inst{2}}
\institute{Huawei Technologies, Munich Research Center, Germany\\\email{\{firstname.lastname\}@huawei.com}\and
Universitat Pompeu Fabra, Music Technology Group, Spain\\
\email{\{firstname.lastname\}01@estudiant.upf.edu}, \email{xavier.serra@upf.edu}}

\begin{document}

\maketitle
\begin{abstract}

Multi-modal deep learning techniques for matching free-form text with music have shown promising results in the field of Music Information Retrieval (MIR).
Prior work is often based on large proprietary data while publicly available datasets are few and small in size. 
In this study, we present \emph{WikiMuTe}, a new and open dataset containing rich semantic descriptions of music.
The data is sourced from Wikipedia's rich catalogue of articles covering musical works.
Using a dedicated text-mining pipeline, we extract both long and short-form descriptions covering a wide range of topics related to music content such as genre, style, mood, instrumentation, and tempo.
To show the use of this data, we train a model that jointly learns text and audio representations and performs cross-modal retrieval.
The model is evaluated on two tasks: tag-based music retrieval and music auto-tagging.
The results show that while our approach has state-of-the-art performance on multiple tasks, but still observe a difference in performance depending on the data used for training.

\keywords{cross-modal \and text-mining \and music information retrieval.}

\end{abstract}
\section{Introduction}\label{sec:introduction}
Music is a complex and multi-faceted art form.
Descriptions of music can therefore be very diverse, covering not only content-related aspects such as instrumentation, lyrics, mood, or other music-theoretical attributes, but also contextual information such as artist or genre-related aspects.
With unprecedented access to vast amounts of music recordings across a multitude of genres and styles, the ability to search for music based on free-form descriptions has become increasingly important in recent years.
Multi-modal deep learning techniques that match textual descriptions with music recordings have seen growing interest \cite{huang_mulan_2022, manco_contrastive_2022, doh_lp-musiccaps_2023,doh_toward_2023,huang_noise2music_2023,mckee_language-guided_2023}.
These methods aim to draw connections between semantic textual descriptions and the content of a music recording and have potential applications in cross-modal retrieval, such as text-to-music search, and automatic music generation.
However, despite promising results, this area of research is still in its infancy, and the lack of suitable and open text-music datasets hinders progress in this field.

MIR researchers have long shown interest in the language metadata for music and have used it to create semantic descriptors.
For example, the use of short text labels (tags) in automatic tagging systems has been studied extensively \cite{bertin-mahieux_automatic_2011,nam_deep_2019}.
These labels are typically taken from online platforms, so-called \textit{social-tags} \cite{lamere_social_2008}.
However, due to data sparsity, researchers often only use a limited number of tags (e.g., top 50), which can cover only certain aspects of the music. 
This highlights the need for a more complex form of description that can capture the nuances of music and provide a richer set of labels for music retrieval and analysis.

To obtain free-form textual descriptions of music, researchers have sought other forms of supervision beyond social-tags.
Some studies have used metadata from online music videos \cite{huang_mulan_2022} or production music \cite{manco_contrastive_2022}.
Since these datasets are not publicly available, different initiatives have proposed to create music description texts manually, either through crowd-sourcing \cite{manco_song_2022} or through expert labelling \cite{agostinelli_musiclm_2023}. 
While both methods can be slow and require significant resources, others have proposed to synthesise text descriptions.
For example, by combining tags into longer text \cite{doh_toward_2023} or by employing large-language models (LLMs) for automatic generation \cite{doh_lp-musiccaps_2023,huang_noise2music_2023,mckee_language-guided_2023}.
While this is a cost-effective method to generate large text corpora, it brings the risk that the LLM produces inaccurate descriptions.

Research on multi-modal training in other areas such as computer vision or machine listening has benefited greatly from the availability of large-scale datasets.
These datasets are often built by crawling the web for suitable data (e.g., images \cite{qi_imagebert_2020,srinivasan_wit_2021} or sound recordings  \cite{wu_large-scale_2023}) paired with natural language texts and may contain millions of data points.
Even though web mining to collect tags for music is common practice in MIR research \cite{turnbull_five_2008}, efforts for text-music descriptions are largely missing to date.
Previous efforts to build multi-modal (music and text) datasets from online sources such as MuMu \cite{oramas_multimodal_2018} or MusiClef2012 \cite{schedl_professionally_2013} collected data at the artist or album level.
We are, however, interested in descriptions of music at a more granular level (i.e., song or segment level).

From these various forms of data sourcing, we identify that 
\begin{enumerate*}[label=(\roman*), itemjoin={{, }}, itemjoin*={{, and }}]
    \item there is a lack of openly accessible datasets
    \item manual labelling is laborious and cannot easily scale to large datasets
    \item synthesised texts might not reflect the resourcefulness of human descriptions
\end{enumerate*}.

To fill this gap, we present \emph{WikiMuTe}, a new publicly available\footnote{\url{https://doi.org/10.5281/zenodo.10223363}} web-sourced dataset of textual descriptions for music collected from encyclopedic articles in Wikipedia.
Using a text-mining pipeline, we extract semantically and syntactically rich text snippets from the articles and associate those with corresponding audio samples.
A cross-modal filtering step is subsequently employed to remove less relevant text-audio pairs.
This approach allows us to collect a large amount of data
describing music content (e.g., genre, style, mood, instrumentation, or tempo) which can serve as a suitable dataset for training deep learning models for text-to-audio matching.
The collected data was utilised in a series of experiments to demonstrate its practical value. 
These experiments involved text-to-music retrieval, music classification, and auto-tagging.
The results are competitive, underlining the value of web-sourced data in these scenarios.

\section{Dataset compilation}

\begin{figure}[t]
    \centering
    \includegraphics[width=0.84\linewidth]{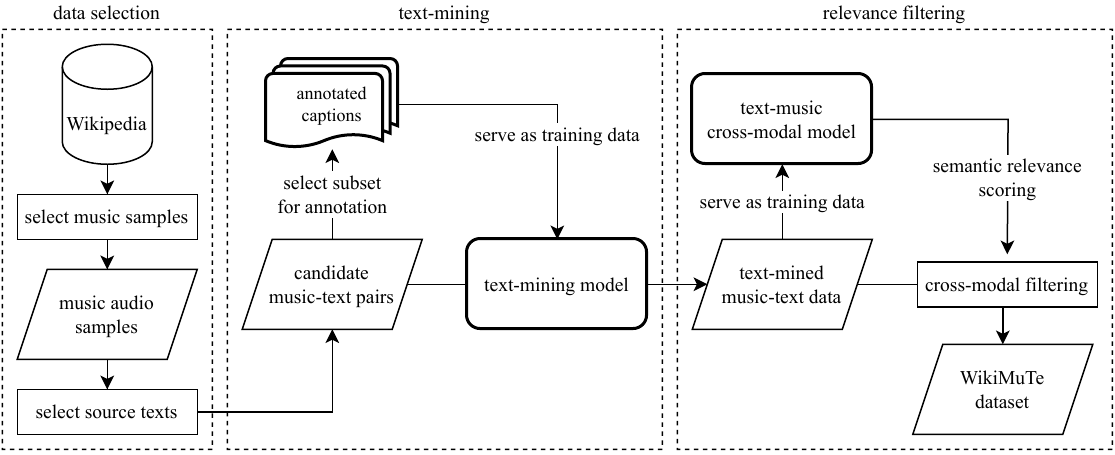}
    \caption{Flowchart of our data mining pipeline}
    \label{fig:pipeline}
\end{figure}

We have chosen the online encyclopedia Wikipedia as our source of data since it contains a vast array of articles on music pieces and popular songs and all content is freely accessible.
To extract a high-quality dataset of music audio paired with textual descriptions, we propose a web mining pipeline consisting of three stages. 
First, we collect music samples and corresponding text from the online encyclopedia Wikipedia.
Since the text covers a wide variety of information about the sample, we extract individual aspects and sentences that refer to descriptions of the music content thereby discarding non-content-related parts.
This is achieved semi-automatically through a dedicated text-mining system.
In the third and final step, we rate the semantic relevance of all music-text pairs and discard texts with low relevance from the final dataset.
Fig. \ref{fig:pipeline} gives an overview of the steps involved.
In the following sections, each of these three stages -- data selection, text-mining and relevance filtering -- is explained in more detail.

\subsection{Source data selection}
\begin{figure}
\centering
 \includegraphics[width=0.6\columnwidth]{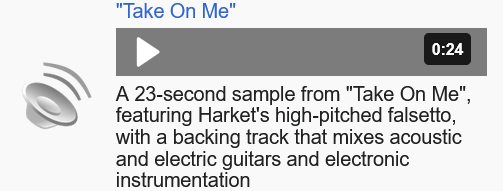}
 \caption{Example of an audio sample with a caption on a Wikipedia page \cite{wikipedia_contributors_take_2023}.}
 \label{fig:audio-sample}
\end{figure}

Some Wikipedia articles include audio samples of music recordings.
These audio samples are often paired with a short text that serves as a caption. 
An example of such a sample and caption is given in Fig.~\ref{fig:audio-sample}.
We selecte audio samples that are linked to the `Music' category on Wikipedia and Wikimedia Commons, effectively excluding samples of speech, natural sounds, etc.
Next, to get the potential text sources, we consider caption texts, file descriptions, and article texts that embed an audio sample if they are linked to Wikipedia's `Songs' category.
To avoid false matches, we exclude certain sections from the article text, such as `Music video', `Chart performance', `Covers' or `Remixes'.
Moreover, we extract categorical information such as genre or instrument names from the metadata if available using simple heuristics.

The collected texts are not readily usable in their raw form.
They frequently contain information that is not pertinent to the audio content, thus introducing irrelevant data that can potentially skew the results of any analysis or application.
Moreover, they present a challenge due to their excessive length.

\subsection{Text-mining system}
To automatically extract music descriptions from the larger bodies of text collected in the first stage, we construct a text-mining system.
This system should be able to handle the complexity of music descriptions and accurately extract relevant information.
Pretrained Transformer models are a state-of-the-art solution for this task and have shown promising results in text mining for other fields \cite{gruetzemacher_deep_2022}.
However, we need to fine-tune them to extract music descriptions.

\subsubsection{Data annotation}
To obtain training data for our text-mining system, we manually labelled the captions of a subset of the previously collected items.
Specifically, we have selected the caption text of the associated audio sample as our annotation source. Caption texts are short and likely to contain descriptions of the music recording.
In our annotations, we focus on descriptions of the musical content only.
Context-dependent descriptions on the other hand, such as lyrics, forms of appraisal, and references to other artists or recordings are deliberately excluded as they cannot easily be matched with the content of the music sample.
We annotated spans of the text in two steps: 
First, we annotated longer phrases such as full sentences and clauses, which we refer to as \textit{sentences}.
Second, we annotated short phrases and single words such as adjectives, nouns, and verbs, which we refer to as \textit{aspects}. 
The former may be described as music captions or long-form descriptions, and the latter as tags or short-form.
The two types of annotations may overlap, as aspects are often part of a longer phrase.%
\footnote{For example, in the caption given in Fig.~\ref{fig:audio-sample} we consider `a backing track that mixes acoustic and electric guitars and electronic instrumentation' a \emph{sentence} and `acoustic and electric guitars' as well as `electronic instrumentation' \emph{aspects}.}

\subsubsection{Text mining \& data collection}\label{subsec:textmining}
Using the annotated data, we train a system to detect sentences and aspects, respectively.
Both systems are trained by fine-tuning a pretrained DistilBERT \cite{sanh_distilbert_2019,devlin_bert_2019} model on a binary token classification task.
This results in automatically extracted spans of text referring to aspects or sentences.
We note that this extracted data is still quite noisy.
It can contain descriptions that are not relevant to the music sample, for example, if the audio represents only a specific part of the song.

\begin{table}[t]
    \centering
    \caption{Two examples of the collected data (RF = Relevance filtering)}
    \begin{tabular}{>{\raggedright}p{2.3cm}p{9.7cm}}
        \toprule
        \multicolumn{2}{c}{Dave Brubeck Quartet - Take Five}\\
       \midrule
        Text-mined aspects & `West Coast cool jazz', `inventive', `cool-jazz', `quintuple', `piano', `saxophone', `bass vamp', `jazz', `two-chord', `roots reggae', `jolting drum solo', `scat'\\
        Text-mined sentences & `is known for its distinctive two-chord piano/bass vamp; catchy, cool-jazz saxophone melodies; inventive, jolting drum solo; and unorthodox quintuple', `the song has a moderately fast tempo of 176 beats per minute', `is a jazz standard'\\
        Removed by RF & `jolting drum solo', `piano', `roots reggae', `the song has a moderately fast tempo of 176 beats per minute'\\
       \midrule
       \multicolumn{2}{c}{Pharrell Williams - Happy}\\
        \midrule
        Text-mined aspects & `handclaps', `faux-', `falsetto voice', `lead vocals', `Soul', `programmed drums', `neo soul', `sings', `uptempo', `156 beats per minute', `neosoul funk groove', `Motown', `falsetto', `backing vocals', `prightly', `mid-tempo', `bass', `sparse', `singing', `vocal', `soul', `keyboard part' \\
        Text-mined sentences & `The song is written in the key of F minor', `and at a tempo of 156 beats per minute', `is an uptempo soul and neo soul song on which', `with a "sprightly neosoul funk groove"', `falsetto voice', `Williams sings the upper notes in falsetto', `is "a mid-tempo ... song in a faux-Motown style, with an arrangement that is, by modern standards, very sparse: programmed drums, one bass and one keyboard part, and handclaps both programmed and played, all topped off by Williams's lead vocals and a whole posse of backing vocals"', `is a mid-tempo soul and neo soul song'\\
        Removed by RF & `sparse', `faux-', `uptempo', `prightly'; `and at a tempo of 156 beats per minute'\\
        \bottomrule
    \end{tabular}
    \label{tab:examples}
\end{table}

\subsection{Semantic relevance filtering}
To ensure that audio and text are matching, we experiment with cross-modal filtering.
Prior work has applied this form of filtering using multi-modal deep learning models as a means to select suitable training data \cite{huang_noise2music_2023,qi_imagebert_2020}. 
By employing models that are trained to give a high score to semantically related data points and a low score to non-matching data points, irrelevant text data can be filtered.
We adopt this idea and employ a model for text-to-audio alignment as a scoring function of how well a description matches a music piece.

To test if the text-mined data are a fitting description for their respective audio sample, we pair each text item with its respective audio sample and compute their score.
For each audio sample that exceeds the input length expected by the model, we compute multiple scores by segmenting the sample into non-overlapping blocks.
The final relevance score is computed by taking the average.
In our implementation, the score is provided as the cosine similarity between the vector representations (embeddings) of audio and text.
We remove all text items with a negative cosine similarity value.

\subsection{The WikiMuTe dataset}

\begin{table*}[t]
\small
\caption{Statistics of text-music datasets}
\begin{tabular}{@{}lr>{\raggedright}p{2.5cm}lp{1.3cm}p{1cm}p{1cm}@{}}
\toprule
Dataset                                     & Tracks                & Item duration                     & Text source   & Vocab. size   & Public text & Public audio \\ \midrule
WikiMuTe                                    & \qty{9}{\thousand}    & snippets \&  full-length tracks   & text-mining   & \num{14392}   & \cmark      & \cmark       \\
MusicCaps \cite{agostinelli_musiclm_2023}   & \qty{5.5}{\thousand}  & \qty{10}{\second}                 & hand-labelled & \num{7578}    & \cmark      & \xmark       \\
LP-MusicCaps \cite{doh_lp-musiccaps_2023}   & \qty{500}{\thousand}  & \qty{30}{\second}                 & LLM generated &  \num{39756}  & \cmark      & \xmark       \\
\addlinespace
eCALS \cite{doh_toward_2023}                & \qty{500}{\thousand}  & \qty{30}{\second}                 & social tags   & \num{1054}    & \cmark      & \xmark       \\
MusCall \cite{manco_contrastive_2022}       & \qty{250}{\thousand}  & full-length tracks                & metadata      & -             & \xmark      & \xmark       \\
MULAN \cite{huang_mulan_2022}               & \qty{40}{\million}    & \qty{10}{\second}                 & metadata      & -             & \xmark      & \xmark       \\ \bottomrule
\end{tabular}
\label{tab:datasets}
\end{table*}
The final dataset contains free-form text descriptions for \num{9000} audio recordings.
On average the tracks have a duration of 81 seconds (Median: \qty{29}{\second}).
The descriptions cover a wide range of topics related to music content such as genre, style, mood, instrumentation, and tempo.
Also, they can include information on the era, production style, and even individual sounds.
Table~\ref{tab:examples} gives two examples of the extracted texts.
Due to its richness, we posit that the data can be used as a  general-purpose music-text dataset and is applicable to several use cases such as cross-modal retrieval and music generation.

Table~\ref{tab:datasets} compares  WikiMuTe with other datasets used in contrastive learning studies for text-to-music matching.
Out of these datasets, the one that stands out as most relevant for comparison is \emph{MusicCaps} \cite{agostinelli_musiclm_2023} due to its size and content, which are comparable to our dataset. 
MusicCaps is derived from the AudioSet dataset \cite{gemmeke_audio_2017} and contains \num{5500} ten-second audio clips labelled with multiple descriptions in the form of a \textit{caption} and an \textit{aspect} list.
It was constructed for music generation from text prompts.
We list the top ten most common aspects in Table~\ref{tab:top_aspects}.
On average, WikiMuTe contains 7.6 (Median: 5) aspects per recording and MusicCaps 10.7 (Median: 9); respectively there are 2.5 (Median: 1) and 3.9 (Median: 4) sentences.

Despite aiming for the inclusion of all kinds of music descriptions, our data can only cover a part of the complexity of music.
In our study, we limit ourselves to the English Wikipedia only since it is the largest.
Other languages are out of the scope of our work.
Given its status as an online encyclopedia, the content in Wikipedia is dynamic and constantly changing.
To get a stable representation we use a database dump of the date 20 July 2023.

\begin{table}[ht]
\centering
\caption{Top ten most common aspects in the datasets used in this study}
\begin{tabular}{l>{\raggedright\arraybackslash}p{10cm}}
\toprule
Dataset   & Top ten aspects           \\ \midrule
WikiMuTe  & classical music, vocals, pop, piano, romantic period classical music, singing, R\&B, sings, vocal, piano music\\
\addlinespace
MusicCaps & low quality, instrumental, emotional, noisy, medium tempo, passionate, energetic, amateur recording, live performance, slow tempo \\
\bottomrule
\end{tabular}
\label{tab:top_aspects}
\end{table}

\section{Experiments}\label{sec:experiments}
To investigate the potential of the collected free-form text data, we experiment with a multi-modal deep learning model for text-to-audio matching.
Such a model maps both modalities, text and audio, into the same embedding space to enable comparisons between the modalities.
To allow for a quantitative evaluation, we select two downstream tasks: tag-based music retrieval and music auto-tagging.
For comparison, we also use the MusicCaps dataset as training data.
In the following section, we provide details about the employed model architecture and explain the downstream task evaluation in Sec.~\ref{sec:evaluation}.

\subsection{System overview \& Implementation details}
\begin{figure}[th]
    \centering
\begin{tikzpicture}[node distance=0.75cm, on grid, every node/.style={font=\small}]
    \node (audio) [rectangle, draw, align=center, text width=1cm] {Audio};
    \node (text) [rectangle, draw, align=center, right=6cm of audio, text width=1cm] {Text};
    \node (encoder_a) [rectangle, draw, text width=1.5cm, align=center, above= of audio, ] {Encoder};
    \node (encoder_t) [rectangle, draw, text width=1.5cm, align=center, above= of text, ] {Encoder};
    \node (adapter_a) [rectangle, draw, text width=1.5cm, align=center, above= of encoder_a, ] {Adapter};
    \node (adapter_t) [rectangle, draw, text width=1.5cm, align=center, above= of encoder_t, ] {Adapter};
    \node (embedding_a) [rectangle, draw,rounded corners, text width=1.6cm, align=center, above of=adapter_a, ] {Embedding};
    \node (embedding_t) [rectangle, draw, rounded corners, text width=1.6cm, align=center, above of=adapter_t, ] {Embedding};
    \node (loss) [rectangle, draw, text width=2.5cm, align=center, right=3cm of embedding_a] {NT-Xent Loss};
    \node[draw,dashed, fit=(encoder_a) (adapter_a)] (tower_a) {};
    \node[draw,dashed, fit=(encoder_t) (adapter_t)] (tower_t) {};
    \draw [->] (audio) -- (encoder_a);
    \draw [->] (text) -- (encoder_t);
    \draw [->] (encoder_a) -- (adapter_a);
    \draw [->] (encoder_t) -- (adapter_t);
    \draw [->] (adapter_a) -- (embedding_a);
    \draw [->] (adapter_t) -- (embedding_t);
    \draw [->] (loss) -- (embedding_a);
    \draw [->] (loss) -- (embedding_t);

    \node (textt_label) [right=2cm of tower_t] {Text tower};
    \node (audiot_label) [left=2cm of tower_a] {Audio tower};
\end{tikzpicture}
    \caption{System overview}
    \label{fig:system}

\end{figure}
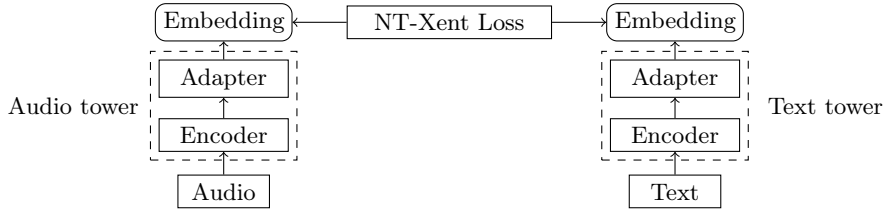

For our system, we adopt a bi-encoder model architecture with an audio tower and a text tower.
Each tower consists of an encoder and a shallow multi-layer perceptron (MLP) adapter.
The purpose of the MLP adapters is to map the different output dimensionalities of the audio and text encoders to a common embedding space.
Fig. \ref{fig:system} shows an overview of the system architecture.
This architecture allows for joint representation learning of both modalities and easily facilitates cross-modal retrieval since each input modality can be encoded separately.
We refer to our system as Music Description Transformer (MDT).

We use the Normalized Temperature-scaled Cross Entropy (NT-Xent) loss \cite{sohn_improved_2016} -- a variant of the InfoNCE loss -- to train our bi-encoder model.
This loss encourages the embeddings of positive pairs (i.e., matching text and audio samples) to be similar while pushing the embeddings of negative pairs (i.e., non-matching text and audio samples) apart. 
For a batch of size $N$ the loss $\mathcal{L}$ is defined as:
\begin{equation}
\mathcal{L}_{NT{\text -}Xent} = \sum_{i=1}^{N} \log \frac{\exp{(s_{i,i} / \tau)}}{\sum_{j=1}^{N} \exp{(s_{i,j} / \tau)}},
\end{equation}
where $\tau$ is a temperature hyperparameter, and $s_{i,j}$ is the cosine similarity between the $i$-th and $j$-th samples in the batch.

For the encoder of the audio tower, we use the \textit{MusicTaggingTransformer} \cite{won_semi-supervised_2021}.
The authors provide a pretrained model, trained on an autotagging task using the Million Song Dataset (MSD) \cite{bertin-mahieux_million_2011}.
The first item of the embedding sequence -- referred to as the \textit{CLS} embedding -- is used as the encoded audio embedding.
The audio encoder expects inputs with a length of \qty{10}{\second}.

For the text encoder, we use a pretrained \textit{SentenceTransformers} language model called \texttt{all-MiniLM-L12-v2} \cite{reimers_sentence-bert_2019, wang_minilm_2020}.
This model is produced using knowledge distillation techniques and gives comparable results to other large language models such as BERT \cite{devlin_bert_2019}, but with fewer model parameters and faster inference.
To extract the text embeddings, we apply mean pooling on the output sequence.
Both adapters of the two tower employ two-layer MLPs with a layer size of 128 and rectified linear units (ReLU) as activation function after the first layer.

To construct a mini batch of audio-text pairs, we randomly pair an audio with a text from its associated \textit{aspects} or \textit{sentences}.
Audio clips in MusicCaps are all 10 seconds long, and can be passed to the audio tower as is.
For WikiMuTe we randomly select a \qty{10}{\second} section from the audio clip.
To get a text from the aspects list we sample up to five aspects at random in both datasets.
We apply the same procedure when selecting a text from the sentences in WikiMuTe.
Since sentences in MusicCaps often form a coherent paragraph, we select $n$ consecutive sentences starting with the first from a sentence list with length $S$, where $n \in [1\ldots S]$.
We use a batch size of 64 and set the temperature parameter $\tau$ to 0.07.

We track the mAP@10 metric on the validation set and stop the training if it does not improve for ten epochs.
The learning rate is set to a initial value of \num{0.0001} and divided by a factor of 10 if no improvement was found for five epochs.
Finally, after completion of the training, the model weights are reverted to the checkpoint of the epoch with the highest score.

\subsection{Experiment settings \& Downstream task evaluation}
\label{sec:evaluation}
To better compare our dataset to previously published datasets, we include MusicCaps as an additional training dataset.
This means we consider two different training datasets: WikiMuTe and MusicCaps.
We test three different configurations of WikiMuTe.
First, using the text-mined data without any cross-modal relevance filtering, i.e. \emph{no filtering} (WMT\textsubscript{nf}).
Second, a \emph{self-filtered} version where the relevance scoring is done by a model trained with the text-mined data (WMT\textsubscript{sf}).
Third, a version filtered by a system trained on the \emph{MusicCaps} training split (WMT\textsubscript{mc}).

\subsubsection{Text-to-music retrieval}
We want to evaluate our system in a setting that best resembles a real-world music search scenario. 
Assuming that users would input short queries rather than full sentences into a text-to-music search system, we propose to adopt an aspect-based (or tag-based) retrieval task \cite{won_multimodal_2021}.
This means we use each aspect/tag in the evaluation data as a query string and perform retrieval over the audio tracks in the test set.
A retrieved track counts as a relevant result if it is labelled with the query string.

To evaluate the effectiveness of our approach we utilise the MusicCaps dataset.
The official evaluation split contains a subset of \num{1000} examples that is balanced with respect to musical genres.
This balanced subset is set aside as the final test data, while the rest of the evaluation data is used as a validation set for model training.
There are \num{4000} distinct aspect strings in the test dataset.
Despite being manually labelled, the dataset contains errors such as typos (e.g., `fats tempo', `big bend').
We do not process the text and leave it for future work to clean the data and possibly merge similar labels (e.g., `no voices', `no vocals').

To measure the performance on these data, we rely on standard information retrieval metrics computed at different numbers of retrieved documents.
Specifically, we use recall at k (R@k) and mean average precision at k (mAP@k) where k denotes the number of results returned for a query, similar to \cite{doh_toward_2023,manco_contrastive_2022}.

\subsubsection{Zero-shot music classification \& auto-tagging}
Additionally, we perform an evaluation in a series of down-stream music classification and tagging tasks.
This allows us to compare our approach to other systems by using several well-established benchmark datasets.
We evaluate in a benchmark proposed by \cite{doh_toward_2023} using multiple datasets that can be grouped into four tasks: music auto-tagging, genre classification, instrument tagging, and mood labelling.
Table~\ref{tab:downstream} lists all datasets used for downstream evaluation and we introduce them briefly.

\begin{table}[t]
\centering
\caption{Overview of the datasets used in downstream zero-shot evaluation. The table is adapted from \cite{doh_toward_2023}.}
\begin{tabular}{lcccc}
\toprule
Dataset        & Task       & Tracks Count  & Tag/Class count & Metric \\
\midrule
MTAT\textsubscript{A} & Tagging    & \num{5329}   & 50     & ROC/PR \\
MTAT\textsubscript{B} & Tagging    & \num{4332}   & 50     & ROC/PR \\
MTG-50     & Tagging    & \num{11356}  & 50   & ROC/PR \\
MTG-G          & Genre      & \num{11479}  & 87   & ROC/PR \\
FMA-Small      & Genre      & \num{800}     & 8   & Acc \\
GTZAN          & Genre      & \num{290}     & 10 & Acc \\
MTG-I          & Instrument & \num{5115}   & 40 & ROC/PR \\
MTG-MT         & Mood/Theme & \num{4231}   & 56 & ROC/PR \\
Emotify        & Mood       & \num{80}      & 9  & Acc \\ \bottomrule
\end{tabular}
\label{tab:downstream}
\end{table}

\begin{description}
\item[MTG-Jamendo \cite{bogdanov_mtg-jamendo_2019}] contains full-length audio tracks and labels taken from an online music platform.
The tags are grouped into four sub-categories: \textit{top 50 tags} (MTG-50), \textit{genre} (MTG-G), \textit{instrument} (MTG-I), and \textit{mood/theme} (MTG-MT).
We use the test split \textit{split-0}.
\item[MagnaTagATune \cite{law_evaluation_2009}] contains 29-second music clips, each of them annotated with labels collected in a crowdsourcing game. We use the 50 most popular tags.
Two splits are commonly used\cite{doh_toward_2023,manco_contrastive_2022,manco_learning_2022}.
Similar to \cite{doh_toward_2023}, we give results for both and refer to them as MTAT\textsubscript{A}
and MTAT\textsubscript{B}.
\item[GTZAN \cite{tzanetakis_musical_2002}] is commonly used for genre classification. It contains 30-second music clips labelled with a single genre category. We employ the fault-filtered split of this dataset \cite{sturm_state_2014}.
\item[Free-Music Archive (FMA) \cite{defferrard_fma_2017}] is a large-scale collection of freely available music tracks.
We use the \textit{small} subset, which is balanced by genre and includes 30-second high-quality snippets of music with a single genre annotation.
\item [Emotify \cite{aljanaki_studying_2016}] contains 1-minute music clips from four genres with annotations for induced musical emotion collected through a crowdsourcing game. Each item is annotated with up to three emotional category labels.
\end{description}

Our model is evaluated in a zero-shot setup, which means that it was not optimised for the task.
Instead, the classification prediction is proxied by the similarity between an audio and the text representation of a class label or tag.

As metrics we use the receiver operating characteristic area under the curve (ROC-AUC) and the precision-recall area under the curve (PR-AUC)  for auto-tagging tasks.
These metrics are especially useful in settings where the number of positive examples is much smaller than the number of negative examples.
ROC-AUC measures the ability of the model to discriminate between positive and negative examples at different classification thresholds. 
PR-AUC focuses on the positive class and measures the trade-off between precision and recall at different classification thresholds.
For classification tasks, we use accuracy (Acc).

\section{Results \& Discussion}

\subsection{Experiment results}

\subsubsection{Text-to-music retrieval}
We average results for the text-to-music retrieval evaluation across three randomly initialised training runs and report their mean and standard deviation in Table~\ref{tab:results}.
As a baseline reference, we include results obtained from a pretrained state-of-the-art system (MTR) \cite{doh_toward_2023}.
which was trained in a contrastive learning task on music and tags using BERT as a text encoder.

\begin{table}[t]
\centering
\caption{Retrieval scores as percentages on the MusicCaps balanced subset.}
\begin{tabular}{llp{1.5cm}p{1.5cm}p{1.5cm}p{1.5cm}}
\toprule
Model                       & Training data & mAP@10 & R@01         & R@05          & R@10  \\ \midrule
MTR       & eCALS                       & $2.8 $        & $0.5 $        & $2.2 $        & $4.3 $ \\ \addlinespace
MDT       & WMT\textsubscript{nf}       &  \num{3.1 \pm .3}  &  \num{0.7 \pm .2}  &  \num{2.7 \pm .2}  &  \num{4.8 \pm .5} \\
MDT       & WMT\textsubscript{sf}       &  \num{3.2 \pm .2}  &  \num{0.7 \pm .1}  &  \num{3.0 \pm .4}  &  \num{5.3 \pm .4} \\
MDT       & WMT\textsubscript{mc}       &  \num{3.4 \pm .2}  &  \num{0.7 \pm .1}  &  \num{3.1 \pm .2}  &  \num{5.6 \pm .4} \\
\addlinespace
MDT       & MusicCaps                   &  \num{5.0 \pm .2}  &  \num{1.0 \pm .1}  &  \num{4.8 \pm .3}  &  \num{8.7 \pm .5} \\
\bottomrule
\end{tabular}
\label{tab:results}
\end{table}

\begin{table}[t]
\caption{Results given as percentages across different downstream tasks 
}
\begin{subtable}[th]{1\textwidth}
\centering
\begin{tabular}{lp{7em}p{6em}p{5.5em}p{5.5em}p{5.5em}}
\hline
\multicolumn{2}{c}{System}                                      & \multicolumn{3}{c}{Tagging} & Instrument \\
\cmidrule(lr){1-2} \cmidrule(lr){3-5} \cmidrule(r){6-6}
Model                                 & Training data           & MTAT-A        & MTAT-B        & MTG50         & MTG-I       \\
                                      &                         & ROC/PR        & ROC/PR        & ROC/PR        & ROC/PR      \\ \cmidrule(lr){1-6}
MTR \cite{doh_toward_2023}            & eCALS                   & 78.4 / 21.2   & 78.7 / 25.2   & 76.1 / 23.6 & 60.6 / 11.3 \\
\addlinespace
MDT                                   & WMT\textsubscript{nf}   & 75.0 / 22.1   & 75.6 / 26.6   & 71.5 / 19.7   & 65.2 / 11.5 \\
MDT                                   & WMT\textsubscript{sf}   & 75.3 / 23.4   & 75.8 / 28.0   & 72.1 / 20.3   & 66.0 / 12.1 \\
MDT                                   & WMT\textsubscript{mc}   & 77.9 / 23.5   & 78.5 / 28.3   & 73.4 / 21.0   & 68.8 / 13.7 \\
\addlinespace
MDT                                   & MusicCaps               & 83.1 / 26.9   & 83.5 / 31.7   & 74.8 / 20.2   & 69.6 / 13.7  \\
 \bottomrule
 \end{tabular}
\end{subtable}
\hfill
\begin{subtable}[h]{1\textwidth}
\centering
\begin{tabular}{lp{7em}p{6em}p{4em}p{4em}p{5.5em}l}
                                      &                         & \multicolumn{3}{c}{Genre}     & \multicolumn{2}{c}{Mood/Theme}                     \\
\cmidrule(lr){3-5} \cmidrule(lr){6-7}
                                      &                         & MTG-G         & GTZAN & FMA   & MTG-MT         & Emot \\
                                      &                         & ROC/PR        & Acc   & Acc   & ROC/PR         & Acc  \\\cmidrule(lr){1-7}

MTR \cite{doh_toward_2023}            & eCALS                   & 81.2 / 15.6   & 87.9  & 45.1  & 65.7 / 8.1    & 33.7 \\
\addlinespace
MDT                                   & WMT\textsubscript{nf}   & 77.9 / 12.5   & 76.9  & 37.8  & 62.9 / 6.4    & 21.3 \\
MDT                                   & WMT\textsubscript{sf}   & 78.1 / 12.8   & 78.2  & 37.1  & 63.4 / 6.3    & 15.0 \\
MDT                                   & WMT\textsubscript{mc}   & 79.2 / 13.3   & 75.5  & 37.1  & 63.1 / 6.8    & 15.0 \\
\addlinespace
MDT                                   & MusicCaps               & 75.9 / 11.5   & 75.9  & 37.3  & 65.4 / 7.7    & 22.5 \\
\bottomrule
\end{tabular}%
\end{subtable}
\label{tab:downstream_eval}
\end{table}

From the table, it can be seen that all WMT configurations outperform the MTR baseline which scores lowest (mean mAP@10 score of \numrange{3.1}{3.4}\% and 2.8\%).
The model trained solely on MusicCaps ranks highest.
This is not surprising since it is the only one that was trained with training data from the same dataset as the test data.
Closer inspection of the table shows that relevance filtering consistently improves the results.
Both the self-filtered data (WMT\textsubscript{sf}) and the data filtered by MusicCaps (WMT\textsubscript{mc}) improve the scores compared to the system trained on unfiltered data (WMT\textsubscript{nf}).

\subsubsection{Zero-shot music classification \& auto-tagging}
We take the best-performing model in each configuration from the previous evaluation and apply it in all zero-shot tasks.
Table~\ref{tab:downstream_eval} shows that our method overall achieves competitive results and outperforms the state-of-the-art system on some of the benchmark datasets, e.g. for tagging (MTAT-A: ROC/PR values of $78.4\% / 21.2\%$ and $77.9\% / 23.5\%$) or instrument classification ($60.6\%/11.3\%$ and $68.8\%/13.7\%$).
However, the MTR baseline achieves higher scores for the genre and mood/theme tagging tasks as well as MTG50.
A possible explanation for this is the much smaller vocabulary in combination with the overall larger dataset size in eCALS (see Table~\ref{tab:datasets}).
Finally, we notice a similar trend as before: the cleaned data through relevance filtering mostly leads to better results. 

\subsection{Findings}
The most obvious findings to emerge from  both evaluations is that the relevance filtering stage is beneficial.
This suggests that there is noisy data in the text-mined collection.
A manual inspection of the removed texts revealed that it provides a form of semantic filtering that is not possible in the earlier stages of the pipeline.
For example, some texts describe parts that are not present in the audio sample, such as the intro or outro or a solo of a song.

Another finding that stands out from the results is that the MusicCaps data generally leads to very good results.
The leading scores in the text-to-music retrieval evaluation could be attributed to the fact that the training and test data come from the same data distribution.
When comparing the metrics in the zero-shot tasks (an out-of-distribution evaluation) the difference less evident.

Finally, we find that the WikiMuTe data can be used to achieve competitive results in text-to-music retrieval and some of the tagging classification tasks.
These results suggest that datasets containing free-text descriptions can enable complex forms of music analysis and retrieval. 

Several factors could explain these observations. 
First, despite the relevance filtering, our data is still noisy and does not match the quality level of hand-labelled data.
Second, our data is more sparse, with longer tracks compared to MusicCaps.
This disparity results in an imbalance in the ratio of text to audio data.
Third, WikiMuTe provides richer data, a point underscored by the fact that its vocabulary size is double that of MusicCaps.
Increasing the amount of audio data would enable us to fully utilise the richness of the text data.
Finally, it is also important to note that texts in Wikipedia were not specifically written to provide comprehensive descriptions of the content of music, but rather for a more general purpose.
As a result, objective descriptors such as instrumentation are likely better represented than more subjective labels such as moods.
In contrast, MusicCaps is a dataset specifically created for music-related content texts.
Despite these challenges, we hypothesize that expanding our approach to include more data sources would enhance our results. 
This assumption is supported by prior studies which demonstrated that larger dataset sizes often lead to superior results, even in the presence of noisy texts \cite{manco_learning_2022, huang_mulan_2022}.

\section{Conclusion}
In this article, we present WikiMuTe, a dataset of rich music descriptions consisting of short labels and long phrases text-mined from Wikipedia articles.
We describe how we construct this web-sourced dataset using a three-stage data-mining pipeline.
To show the use of the data, we study how it can be leveraged to fine-tune audio and text models to match music audio with musically-relevant textual descriptions.
We evaluate our trained system in two ways: tag-based music retrieval and music tagging.
While the results achieved with a system trained with the WikiMuTe data are generally better or comparable to those of a state-of-the-art system, we observe that data from a manually created dataset can lead to even higher scores.
Moreover, we find that filtering the text-mined data according to a cross-modal relevance scoring, leads to improved results.
Future work could investigate if expanding the data mining approach to multiple web sources helps to further improve the results.
\bibliographystyle{splncs04}
\bibliography{ref}
\end{document}